\documentclass[10pt,twocolumn,letterpaper]{article}

\usepackage{iccv}
\usepackage{times}
\usepackage{epsfig}
\usepackage{graphicx}
\usepackage{amsmath}
\usepackage{amssymb}
\usepackage{color}

\usepackage[pagebackref=true,breaklinks=true,letterpaper=true,colorlinks,bookmarks=false]{hyperref}

\iccvfinalcopy 

\def\iccvPaperID{7538} 

\ificcvfinal\fi

\newcommand*{\affaddr}[1]{#1} 
\newcommand*{\affmark}[1][*]{\textsuperscript{#1}}
\newcommand*{\email}[1]{\texttt{#1}}

\begin{document}

\title{Half-Real Half-Fake Distillation for Class-Incremental Semantic Segmentation}

\author{%
    Zilong Huang\affmark[1]$^{*}$, Wentian Hao\affmark[2]\thanks{equal contribution},  Xinggang Wang\affmark[1], Mingyuan Tao\affmark[2], Jianqiang Huang\affmark[2], \\ Wenyu Liu\affmark[1], Xian-Sheng Hua\affmark[2] \\
    \affaddr{\affmark[1]School of EIC, Huazhong University of Science and Technology}\\
    \affaddr{\affmark[2]Alibaba Group }\\
    \email{\tt\small zilong.huang2020@gmail.com \{xgwang,liuwy\}@hust.edu.cn } \\ \email{\tt\small\{wentian.hwt,juchen.tmy,jianqiang.jqh\}@alibaba-inc.com huaxiansheng@gmail.com} \\
}

\maketitle
\ificcvfinal\thispagestyle{empty}\fi

\begin{abstract}
    Despite their success for semantic segmentation, convolutional neural networks are ill-equipped for incremental learning, \ie, adapting the original segmentation model as new classes are available but the initial training data is not retained. Actually, they are vulnerable to catastrophic forgetting problem. We try to address this issue by ``inverting" the trained segmentation network to synthesize input images starting from random noise. To avoid setting detailed pixel-wise segmentation maps as the supervision manually, we propose the SegInversion to synthesize images using the image-level labels. To increase the diversity of synthetic images, the Scale-Aware Aggregation module is integrated into SegInversion for controlling the scale (the number of pixels) of synthetic objects. Along with real images of new classes, the synthesized images will be fed into the distillation-based framework to train the new segmentation model which retains the information about previously learned classes, whilst updating the current model to learn the new ones. The proposed method significantly outperforms other incremental learning methods, and obtains the state-of-the-art performance on the PASCAL VOC 2012 and ADE20K datasets.
\end{abstract}

\section{Introduction}

Semantic segmentation, aims at assigning semantic class labels to each pixel in a given image, provides significant impacts on various real-world applications, such as autonomous driving~\cite{fritsch2013new}, augmented reality~\cite{azuma1997survey}, \etc. Specifically, current state-of-the-art semantic segmentation approaches based on the fully convolutional network (FCN)~\cite{long2015fully} have made remarkable progress in several ways, \eg, by modeling context information~\cite{zhao2017pyramid,chen2017rethinking,zhang2018context,wang2018non,huang2019ccnet}, recovering the spatial details~\cite{chen2018encoder,ronneberger2015u,huang2020alignseg} or designing stronger networks~\cite{yu2018deep,wang2020deep,pohlen2017full}. 

However these methods struggle in incrementally learning new
tasks whilst preserving good performance on previous ones when new categories are added. Meanwhile, for protecting the data privacy, the requirement for prior training data can be very restrictive in practice. The methods will suffer from a dramatic decrease in performance on old classes when updating the model without the previous training data (including both training images and labels).


\begin{figure}[!t]
    \centering
    \includegraphics[width=1.0\linewidth]{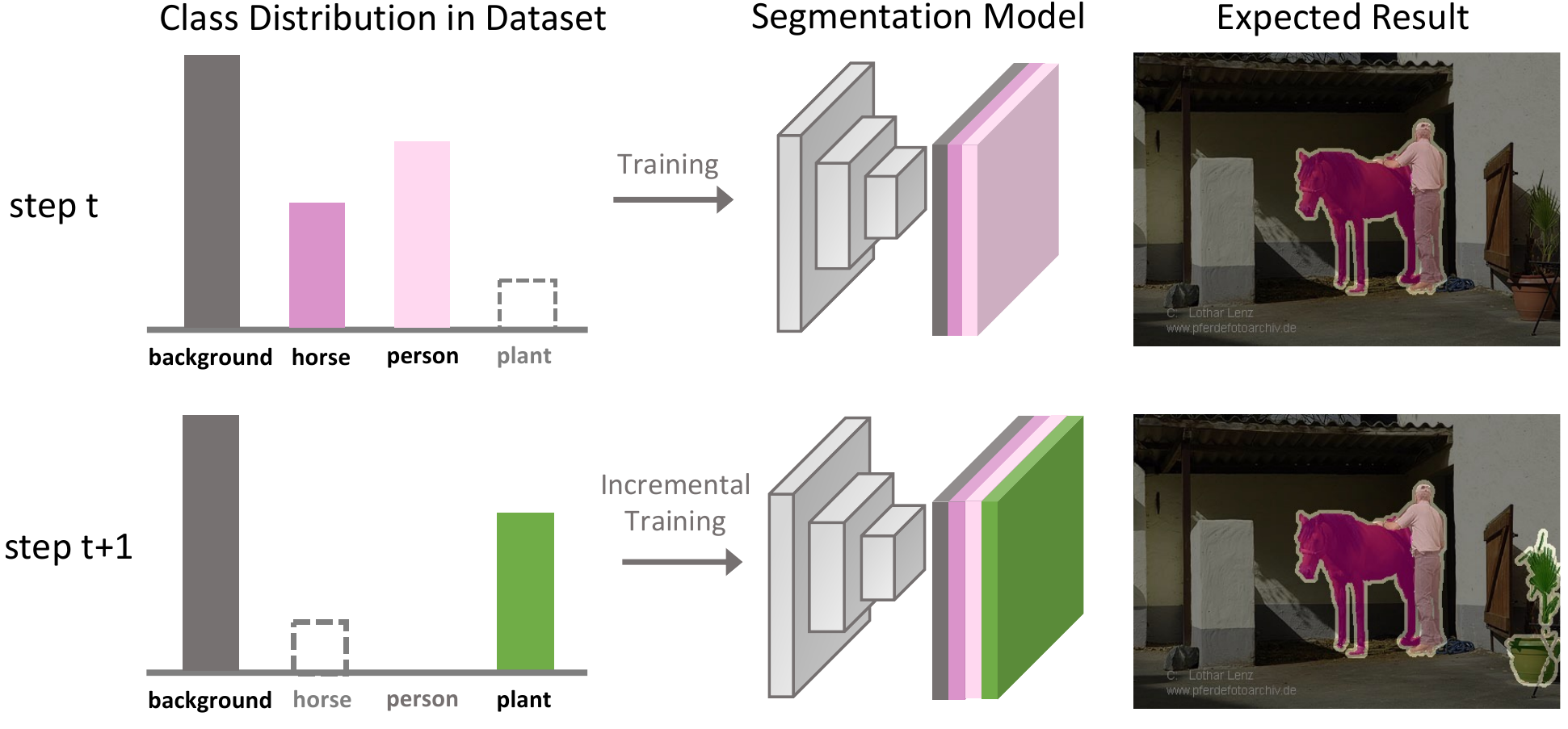}
    \caption{The setting of class-incremental semantic segmentation. Dashed bars in class distribution denote there exist some training samples of the corresponding classes in training images, following the settings of ~\cite{cermelli2020modeling}, they will be labeled as background.}
    \label{fig:task}
    \vspace{-5mm}
\end{figure}

Considering the example in Figure~\ref{fig:task},
a segmentation model is trained for classes background, horse and person in step $t$ and is further trained for classes background, horse, person and plant in step $t+1$ \textbf{only} with the new dataset. In the new dataset, the new class plant is labeled, the others (dashed bars) are labeled as background. After training on the new dataset, the network is expected to segment all four classes. However, the fact is, the network can segments the plant in the image, but may fails to segment the person and horse, which is referred as \textit{catastrophic forgetting}.

While the problem of incremental learning has been traditionally addressed in object recognition~\cite{chaudhry2018riemannian, li2017learning, kirkpatrick2017overcoming} and detection~\cite{shmelkov2017incremental}, much less attention has been devoted to semantic segmentation~\cite{michieli2019incremental,michieli2021continual,cermelli2020modeling}. 
To alleviate the catastrophic forgetting issue of incremental learning in semantic segmentation,  Michieli \etal \cite{michieli2019incremental} utilize knowledge distillation~\cite{hinton2015distilling} frameworks with new designed distillation losses to preserve previously learned knowledge when learning on new data. However, there are two obstacles remaining. 1) Semantic distribution shift within the background class, as shown in Figure~\ref{fig:task}, the class plant is labeled as background in step $t$ but is labeled as foreground in step $t+1$. 2) Lack of samples of old classes, as shown in Figure~\ref{fig:task}, the new dataset does NOT contain samples of the old class person and has a few samples of the old class horse. The distillation will be inefficient if the new 
images can not activate the response of the ``Teacher" model (previous trained model) on the old classes. For the obstacle \#1, Cermelli \etal \cite{cermelli2020modeling} introduce a new objective function to explicitly cope with the evolving semantics of the background class. In this paper, we mainly study the obstacle \#2.

In the absence of prior training data, an interesting question arises – can we somehow generate fake training samples for old classes from the already trained segmentation model and use both fake data of old classes and real data of new classes in distillation framework for incremental learning?  A lot of methods have attempted to synthesizes images from the trained classification model~\cite{yin2020dreaming, mordvintsev2015inceptionism}. The most popular and simple-to-use method is DeepDream~\cite{mordvintsev2015inceptionism}. It synthesizes or transforms an input image to yield high output responses for chosen classes in the output layer of a given classification model. For segmentation segmentation, it is difficult to give a preset detailed segmentation map in output layer of segmentation model. Inspired by the Multiple Instance Learning (MIL) based pooling method~\cite{pinheiro2015image}, we introduce a Scale-Aware Aggregation module (SAA) to adaptively pool the segmentation map into the classification scores. This aggregation module will put more weight on pixels which are important for the classification decision. Thus, we could choose classes manually as the supervision instead of setting detailed pixel-wise segmentation maps. Furthermore, The proposed SAA could adjust the scale (the number of pixels) of objects belonging to pre-assumed classes. It could help to generate more diverse fake data. The fake data generation method for segmentation model is named as SegInversion.

After generating fake data, the fake images and real images of new classes will be fed into a distillation-based framework to train a new segmentation model. The fake data can be used to activate response of the ``Teacher'' model, which makes it efficient for ``Student'' model to learn knowledge of old classes. The real data can supervise ``Student'' model to learn knowledge of new classes. The distillation-based framework of incremental learning is named as Half-Real Half-Fake Distillation (HRHF).

To demonstrate the effectiveness of our approach,
we conduct elaborate experiments on the challenging
PASCAL VOC 2012~\cite{everingham2010pascal} and ADE20K~\cite{zhou2017scene} datasets. Our approach
obtains the state-of-the-art performance under various settings.
To summarize, the contributions of this paper are as follows:
\begin{itemize}
    \item We propose the Half-Real Half-Fake Distillation framework for incremental learning, which use fake data of old classes and real data of new classes simultaneously to update the new segmentation model.
    \item We propose the SegInversion which uses the previous trained segmentation model with the image-level labels to synthesize fake images. The core of SegInversion is the Scale-Aware Aggregation module which can control the scale (the number of pixels) of synthetic objects.
    \item Our work outperforms several previous incremental learning methods by a large margin (\ie, up to ~35\% mIoU on PASCAL VOC 2012) and obtains the state-of-the-art performance on the
    PASCAL VOC 2012 and ADE20K datasets. 
\end{itemize}

\begin{figure*}[!t]
    \centering
    \includegraphics[width=1.0\linewidth]{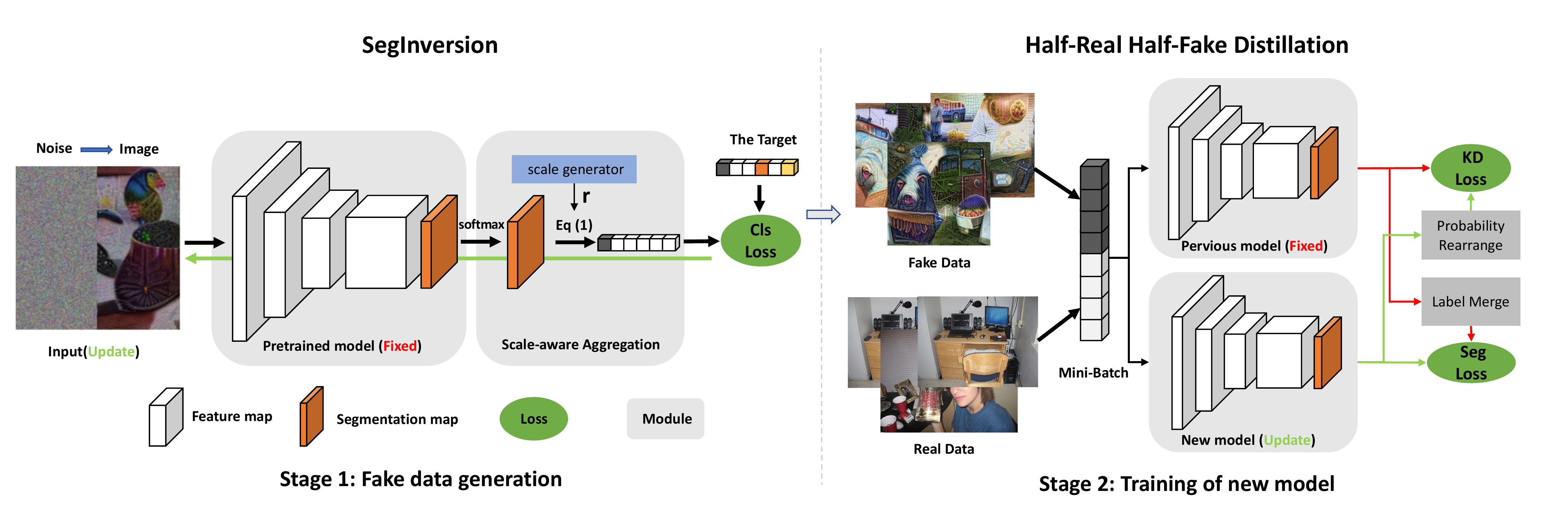}
    \caption{The pipeline of our Half-Real and Half-Fake framework for incremental learning. There are two stages in the pipeline, the SegInversion and Half-Real Half-Fake Distillation. The SegInversion  optimizes random noise into high-fidelity class-conditional images given just a pretrained segmentation model. In SegInversion, the Scale-Aware Aggregation (SAA) is proposed to adaptively aggregate the segmentation maps into classification scores. The random generated $r$ is used in SAA to control the scale of the synthetic object. In Half-Real Half-Fake Distillation, we feed the fake data to preserve old knowledge and real data to learn new classes into the distillation framework. Cooperated with the distillation loss and segmentation loss, the Half-Real Half-Fake Distillation could alleviate the catastrophic forgetting issue of incremental learning in semantic segmentation. 
    }
    \label{fig:architecture}
    \vspace{0mm}
\end{figure*}

\section{Related work} \label{Related work}

\noindent\textbf{Semantic Segmentation}
The last years have seen a renewal of interest on semantic segmentation. FCN~\cite{long2015fully} is the first approach to adopt fully convolutional network for semantic segmentation. Current state-of-the-art semantic segmentation approaches based on the fully convolutional network (FCN)~\cite{long2015fully} have made remarkable progress in several ways, e.g. by modeling context information~\cite{zhao2017pyramid,chen2017rethinking,zhang2018context,wang2018non,huang2019ccnet}, recovering the spatial details~\cite{chen2018encoder,ronneberger2015u,huang2020alignseg} or designing stronger networks~\cite{yu2018deep,wang2020deep,pohlen2017full}. The vast majority of semantic segmentation methods consider a static setting, \ie, the training data for all classes are available before training. There are few works~\cite{michieli2019incremental, ozdemir2018learn, ozdemir2019extending, tasar2019incremental, cermelli2020modeling} that focus on the problem of incremental learning in semantic segmentation. Ozdemir~\etal~\cite{ozdemir2018learn, ozdemir2019extending} designed the incremental learning approach for medical image segmentation and adapt a standard image-level classification method~\cite{li2017learning} to segmentation and devising a strategy to select relevant samples of old datasets for rehearsal. Taras~\etal~\cite{tasar2019incremental} proposed a similar approach for segmenting remote sensing data. Michieli~\etal~\cite{michieli2019incremental} considered incremental learning for semantic segmentation in a particular setting where labels are provided for old classes while learning new ones. Recently, Cermelli~\etal~\cite{cermelli2020modeling} proposed a more principled formulation of the incremental learning problem in semantic segmentation and address the problem of modeling the background shift of the semantic segmentation setting. Both \cite{michieli2019incremental} and \cite{cermelli2020modeling} utilize a distillation framework to transform the knowledge of old classes from the previous segmentation network to the new segmentation network, meanwhile update the new model on the data of new classes. Different from \cite{michieli2019incremental} and \cite{cermelli2020modeling}, the proposed method mainly study the problem of lack of samples of old classes during distillation, which is failed to be considered before.

\noindent\textbf{Incremental Learning}
Incrementally learning several tasks with one neural network exhibits the problem of catastrophic forgetting~\cite{mccloskey1989catastrophic}. The problem has been extensively studied for image classification tasks~\cite{de2019continual,caccia2020online, yu2020semantic}. The works can be grouped into three categories: replay-based~\cite{castro2018end, hou2019learning, ostapenko2019learning, rebuffi2017icarl, shin2017continual, wu2018memory, ebrahimi2020adversarial}, regularization-based~\cite{dhar2019learning, chaudhry2018riemannian, kirkpatrick2017overcoming, li2017learning}, and parameter isolation-based~\cite{mallya2018piggyback, mallya2018packnet, rusu2016progressive}.  In replay-based methods, examples of previous tasks are either stored~\cite{castro2018end, hou2019learning,rebuffi2017icarl} or
generated~\cite{ostapenko2019learning, shin2017continual,wu2018memory} and then replayed while learning the new task. Parameter isolation-based methods~\cite{mallya2018piggyback, mallya2018packnet, rusu2016progressive}
assign a subset of the parameters to each task to prevent forgetting. Regularization-based methods can be divided into
prior-focused~\cite{chaudhry2018riemannian, kirkpatrick2017overcoming} and data-focused~\cite{dhar2019learning, li2017learning}.

The methods~\cite{michieli2019incremental, cermelli2020modeling} for incremental learning in semantic segmentation could also be grouped into regularization-based methods. They utilize a knowledge distillation framework to use the output of previous segmentation as the supervision of new segmentation network.

\noindent\textbf{Image synthesis}
The works could be grouped int two categories: Generative adversarial networks (GANs) based~\cite{goodfellow2014generative, ostapenko2019learning, shin2017continual,wu2018memory} and Model Inversion based~\cite{fredrikson2015model, mordvintsev2015inceptionism, yin2020dreaming}. Generative adversarial networks (GANs)~\cite{goodfellow2014generative} have been at the forefront of generative image modeling, yielding high-fidelity images. Though adept at capturing the image distribution, training a GAN’s generator requires access to the original data.

Fredrikson \etal~\cite{fredrikson2015model} propose the model inversion attack to obtain class images from a network through a gradient descent on the input. And its followup DeepDream~\cite{mordvintsev2015inceptionism} and DeepInversion~\cite{yin2020dreaming} further improves image diversity and quality. Despite notable progress on synthesizing images based on classification network, much less attention has been devoted to synthesize images from segmentation network. Because It's almost impossible to manually generate the 2D segmentation map as the target. The proposed SegInversion with Scale-Aware Aggregation module could make it possible to synthesize images from segmentation network with the image-level labels .

\section{Approach} \label{HRHF approach}
Our Half-Real Half-Fake framework consists of two stages: (i) SegInversion for fake data generation, and (ii) Half-Real Half-Fake Distillation for learning new segmentation model. In this section, we briefly discuss the background and notation, and then introduce the fake data generation and the training of the new segmentation model.

\subsection{Problem Definition and Notation}
We first introduce the task of incremental learning for semantic segmentation. The incremental learning for semantic segmentation can be defined as the ability of a learning system (e.g., a neural network) to learn the segmentation and the labeling of the new classes without forgetting or deteriorating too much the performance on previously learned ones~\cite{michieli2019incremental}.

The training is realized over multiple phases, called $learning\,steps$, and each step $t$ introduces novel categories $\mathcal{C}^t$ to be learnt. The labels of new data are donated as $y^t$. The number of categories to be segmented at step $t$ is donated as $C^t$. We assume that we have trained our network with the parameters $\theta_t$ at step $t$. At learning step $t$ we are also provided with a training set $\mathcal{D}^t$ that is used in conjunction to the previous segmentation model $\theta_{t-1}$ to train and update the model $\theta_t$.   Following the standard incremental learning settings, the sets of labels $\mathcal{C}^t$ at the different learning steps are disjoint, except for the special void/background class.

\subsection{Fake Data Generation with SegInversion}\label{SegInversion}
Inspired by DeepDream~\cite{mordvintsev2015inceptionism} and DeepInversion~\cite{yin2020dreaming}, we proposed the SegInversion to synthesize high-fidelity and high-resolution images. To make it possible to generate fake data using the trained segmentation network with the image-level labels, we introduce the Scale-Aware Aggregation module. This module could adaptively pools the 2D segmentation map into the classification scores.

As shown in Figure~\ref{fig:architecture}, given a randomly initialized input $x \in R^{H \times W \times C}$, $H, W, C$ being the
height, width, and number of color channels. The previous trained segmentation model $\theta_{t-1}$ takes the $x$ as input and output the segmentation map $s \in R^{H_o \times W_o \times C^{t-1}}$, $H_o, W_o$ being the height, width of the output. Considering the objects of some categories have special shapes and contexts, it is too complex to generate an appropriate segmentation map  as the 2D target manually. To address this issue, we need a way to
aggregate these pixel-level scores into a single image-level
classification score $\hat{y} \in R^{C^{t-1}}$, $C^{t-1}$ is the number of categories in the previous step. Thus, we can use an arbitrary
1D one-hot vector $y \in R^{C^{t-1}}$ as the target label. The aggregation should drive the network
towards appropriate pixel-level assignments, such that it could
perform decently on segmentation tasks. 

Inspired by LSE pooling~\cite{pinheiro2015image}, we propose the Scale-Aware Aggregation approach to adaptively adjust the number of pixels belonging to the category.
\begin{align} \label{eq:optimizing}
\hat{y}^k=\frac{1}{r}\log[\frac{1}{{H_o}{W_o}}\sum_{i,j}exp(r*s^k_{i,j})] \quad \forall k \in \mathcal{C}^{t-1}.
\end{align}

The hyper-parameter $r$ controls the object scale belonging to the category: high $r$ values implies having an effect similar to the max aggregation, very low $r$ values will have an effect similar to the score averaging\cite{pinheiro2015image}. This means, high $r$ will promote to generate objects with the small scales, while low $r$ will promote to generate objects with the large scales. Different from~\cite{pinheiro2015image}, $r$ could be randomly generated for each category and each input image. The advantage of this aggregation is to increase the diversity of the fake data.

Assuming an $r$ is chosen randomly and the segmentation map is aggregated into the the classification scores $\hat{y}$. The fake image could be synthesized by optimizing,
\begin{align} \label{eq:optimizing}
\min_{x}\ell_{cls}(\hat{y}, y) + R(x).
\end{align}
where $\ell_{cls}(\cdot)$ is a classification loss (\eg, cross-entropy), and $R(\cdot)$ is an image regularization term. Following DeepInversion~\cite{yin2020dreaming}, we use an image prior~\cite{mordvintsev2015inceptionism} and feature distribution regularization term~\cite{yin2020dreaming} as $R(\cdot)$. In the optimization step, the learnable parameters of trained segmentation are all fixed and errors between the target and the classification scores $\hat{y}$ are back-propagated to $x$ and update $x$ by the gradients. Then, the updated image will be passed into the network repeatedly until the loss is lower than the preset value.

\subsection{Half-Real Half-Fake Distillation for Incremental Learning} \label{distillation}
To better preserve old knowledge in the pretrained model, we choose the distilling framework which uses the prediction of the old model $\theta_{t-1}$ as the supervision of new segmentation network $\theta_t$.
Different from the previous methods, we could use the fake data and real data to drive the distillation framework to update the segmentation network. As shown in Fig.~\ref{fig:architecture}, the images of the mini-batch are sampled from the fake data and real data, then the batched images are fed into the previous (old) trained segmentation network and the student (new) segmentation network to produce segmentation map $s^{t-1} \in R^{H_o \times W_o \times C^{t-1}}$ and $s^{t} \in R^{H_o \times W_o \times C^t}$, respectively.

Here, we introduce a distillation loss to preserve old knowledge and a segmentation loss to learn new classes. Following~\cite{cermelli2020modeling}, we use \textbf{Probability Rearrange} to solve the semantic distribution shift within the background class. The Probability Rearrange module takes $s^{t}$ as input and add probabilities of the new classes into the background, then the probabilities of old classes (including background) are selected to produce $\hat{s}^t \in R^{H_o \times W_o \times C^{t-1}}$. Thus, after probability rearrangement, the adapted $\hat{s}^t$ has the same shape with $s^{t-1}$. We use the distillation loss to encourage the match of two probability maps. The distillation loss is defined as follow:
\begin{align} \label{eq:kd}
\ell_{kd}^t = -\frac{1}{H_oW_o}\sum_u\sum_{c\in\mathcal{C}^{t-1}}s_{u,c}^{t-1}\log{\hat{s}_{u,c}^t}.
\end{align}

In addition to building a distillation loss using the ``soft'' probabilities, we also design a segmentation loss on the ``hard'' index map and segmentation map $s^t$. To obtain complete labels of all classes, the \textbf{Label Merge} module firstly converts the label of new classes into one-hot map, then concatenates the probability map $s^{t-1}$ and the one-hot map, to obtains the index map for all classes by applying argmax function on the concatenated map. Finally, the index map is converted into the one-hot map $y^{merge}$. 
The segmentation loss is defined as follow:

\begin{align} \label{eq:seg}
\ell_{seg}^t = -\frac{1}{H_oW_o}\sum_uy^{merge}_u\log{s_u^t}.
\end{align}

In particular, we minimize a loss function of the following form. $\lambda > 0$ is a hyper-parameter balancing the importance of the two terms.
\begin{align} \label{eq:optimizing}
\ell_{all} = \lambda\ell_{kd} + \ell_{seg}.
\end{align}
\section{Experiments} \label{Experiments}

\begin{table*}[!t]
\renewcommand{\arraystretch}{1.3}
\setlength{\tabcolsep}{0.15em}
\caption{Comparison of mIoU on the PASCAL VOC 2012 dataset with different methods}
\label{tab:pascal}
\centering
    \begin{tabular}{c|ccc|ccc||ccc|ccc||ccc|ccc}
    \hline
    {} & \multicolumn{6}{c||}{\textbf{19-1}} & \multicolumn{6}{c||}{\textbf{15-5}} & \multicolumn{6}{c}{\textbf{15-1}}\\
    {} & \multicolumn{3}{c|}{\textbf{Disjoint}} & \multicolumn{3}{c||}{\textbf{Overlapped}} & \multicolumn{3}{c|}{\textbf{Disjoint}} & \multicolumn{3}{c||}{\textbf{Overlapped}} & \multicolumn{3}{c|}{\textbf{Disjoint}} & \multicolumn{3}{c}{\textbf{Overlapped}}\\
    \textbf{Method} & \textit{1-19} & \textit{20} & \textit{all} & \textit{1-19} & \textit{20} & \textit{all} & \textit{1-15} & \textit{16-20} & \textit{all} & \textit{1-15} & \textit{16-20} & \textit{all} & \textit{1-15} & \textit{16-20} & \textit{all} & \textit{1-15} & \textit{16-20} & \textit{all}\\
    \hline
    FT & 5.8 & 12.3 & 6.2 & 6.8 & 12.9 & 7.1 & 1.1 & 33.6 & 9.2 & 2.1 &  33.1 & 9.8 & 0.2 & 1.8 & 0.6 & 0.2 & 1.8 & 0.6\\
    PI\cite{zenke2017continual} & 5.4 & 14.1 & 5.9 & 7.5 & 14.0 & 7.8 & 1.3 & 34.1 & 9.5 & 1.6 &  33.3 & 9.5 & 0.0 & 1.8 & 0.4 & 0.0 & 1.8 & 0.5\\
    EWC\cite{kirkpatrick2017overcoming} & 23.2 & 16.0 & 22.9 & 26.9 & 14.0 & 26.3 & 26.7 & 37.7 & 29.4 & 24.3 & 33.5 & 27.1 & 0.3 & 4.3 & 1.3 & 0.3 & 4.3 & 1.3\\
    RW\cite{chaudhry2018riemannian} & 19.4 & 15.7 & 19.2 & 23.3 & 14.2 & 22.9 & 17.9 & 36.9 & 22.7 & 16.6 & 34.9 & 21.2 & 0.2 & 5.4 & 1.5 & 0.0 & 5.2 & 1.3\\
    LwF\cite{li2017learning} & 53.0 & 9.1 & 50.8 & 51.2 & 8.5 & 49.1 & 58.4 & 37.4 & 53.1 & 58.9 & 36.6 & 53.3 & 0.8 & 3.6 & 1.5 & 1.0 & 3.9 & 1.8\\
    LwF-MC\cite{rebuffi2017icarl} & 63.0 & 13.2 & 60.5 & 64.4 & 13.3 & 61.9 & 67.2 & 41.2 & 60.7 & 58.1 & 35.0 & 52.3 & 4.5 & 7.0 & 5.2 & 6.4 & 8.4 & 6.9\\
    ILT\cite{michieli2019incremental} & 69.1 & 16.4 & 66.4 & 67.1 & 12.3 & 64.4 & 63.2 & 39.5 & 57.3 & 66.3 & 40.6 & 59.9 & 3.7 & 5.7 & 4.2 & 4.9 & 7.8 & 5.7\\
    MiB\cite{cermelli2020modeling} & 69.6 & 25.6 & 67.4 & 70.2 & 22.1 & 67.8 & 71.8 & 43.3 & 64.7 & 75.5 & 49.4 & 69.0 & 46.2 & 12.9 & 37.9 & 35.1 & 13.5 & 29.7\\
    HRHF & \textbf{76.8} & \textbf{56.3} & \textbf{75.8} & \textbf{76.6} & \textbf{57.3} & \textbf{75.7} & \textbf{76.2} & \textbf{49.2} & \textbf{69.5} & \textbf{78.9} & \textbf{57.8} & \textbf{73.9} & \textbf{72.6} & \textbf{32.2} & \textbf{62.5} & \textbf{72.4} & \textbf{39.6} & \textbf{64.6}\\
    \hline
    Joint & 77.4 & 48.0 & 77.4 & 77.4 & 78.0 & 77.4 & 79.1 & 72.6 & 77.4 & 79.1 & 72.6 & 77.4 & 79.1 & 72.6 & 77.4 & 79.1 & 72.6 & 77.4\\
    \hline
    \end{tabular}
    \vspace{-4mm}
\end{table*}

\subsection{Experimental setup}


We give a comprehensive comparison between our method and the previous state-of-the-art method MiB~\cite{cermelli2020modeling} on commonly used benchmarks for semantic segmentation: PASCAL VOC 2012~\cite{everingham2010pascal} and ADE20K~\cite{zhou2017scene}. We employ the same experimental setups in both datasets as in MiB~\cite{cermelli2020modeling} and prove that our method achieves new state-of-the-art results. All results are reported as mean Intersection-over-Union (mIoU) in percentage averaged over all the classes. Two more baselines are listed as in MiB~\cite{cermelli2020modeling}: simple fine-tuning (FT) in each step and training on all classes off-line (Joint). The latter one can be regarded as an upper bound in performance. Besides, we also list other incremental learning methods for comparison in Table~\ref{tab:pascal} and Table~\ref{tab:ade} for reference: incremental learning methods like Elastic Weight Consolidation(EWC)~\cite{kirkpatrick2017overcoming}, Path Integral(PI)~\cite{zenke2017continual} and Riemannian Walks(RW)~\cite{chaudhry2018riemannian}, which all employ different strategies to compute the importance of parameters for old classes, and also data-focused methods like Learning without forgetting(LwF)~\cite{li2017learning}, LwF multi-class(LwF-MC)~\cite{rebuffi2017icarl} and the segmentation method ILT~\cite{michieli2019incremental}.

\subsection{Implementation Details}

For semantic segmentation, as in MiB\cite{cermelli2020modeling}, we use the same Deeplab-v3 architecture\cite{chen2017rethinking} with a ResNet-101\cite{he2016deep} backbone and output stride of 16. We also use in-place activated batch normalization as proposed in \cite{rota2018place} and apply the same data augmentation and training hyper-parameters as in \cite{cermelli2020modeling}. 

For fake data generation, we use Adam optimizer as in \cite{bhardwaj2019dream} with learning rate 0.25. The batch size is 16 and the resolution of input noise is 512 x 512. The training step is set to 1000.

The $\lambda$ is set to 1 for balancing the distillation loss and the segmentation loss.

\subsection{PASCAL VOC 2012}

PASCAL VOC 2012~\cite{everingham2010pascal} is one of the widely used benchmarks for semantic segmentation, which has 20 foreground object classes. Following~\cite{cermelli2020modeling,michieli2019incremental,shmelkov2017incremental}, we adopt two experimental settings depending on how images are sampled while building the incremental datasets: $disjoint$ and $overlapped$. For the first setup, each learning step contains a unique set of images, whose pixels belong to classes seen either in the current or in the previous steps. For the second setup, each learning step contains all the images that have at least one pixel of a novel class, which means that images may now contain classes that will be learned in the future. In both setups for each incremental learning step, there are only labels for novel classes, while other classes are labeled as background in the ground truth. We report results for three different experiments as in \cite{cermelli2020modeling,michieli2019incremental,shmelkov2017incremental}: the addition of one class ($19$-$1$), five classes all at once ($15$-$5$) and five classes sequentially ($15$-$1$).\\


\noindent\textbf{Addition of one class ($19$-$1$).} In this setting, two learning steps are adopted: learning 19 classes in the first step and the last $tv$-$monitor$ class in the second step. Results are reported in Table~\ref{tab:pascal}. Performance with simple fine-tuning drops significantly. Our method surpasses others in performance by a large margin in both $disjoint$ and $overlapped$ setups. To be more specific, our method alleviates the catastrophic forgetting successfully in the second step while keeping a much better performance for the new class and improves 8$\%$ in overall mIoU compared with MiB\cite{cermelli2020modeling}, which has the best performance among previous methods. \\


\noindent\textbf{Single-step addition of five classes ($15$-$5$).} In this experiment, 15 classes are observed by model during training in the first step and the left 5 classes ($plant$, $sheep$, $sofa$, $train$ and $tv$-$monitor$) are added simultaneously in the second step. Our method still surpasses all the other methods in all steps and gains 5$\%$ in overall mIoU compared with MiB~\cite{cermelli2020modeling}.\\


\noindent\textbf{Multi-step addition of five classes ($15$-$5$).} This setting is similar to the previous one except that the last 5 classes are learned sequentially one by one in 5 steps. This is a more difficult tasks than the last two experiments, and all the methods almost totally failed in this scenario except MiB~\cite{cermelli2020modeling}. Our method further outperforms it by a large margin in both old and new classes. Overall, our method improves the performance in mIoU in this scenario by 24.6$\%$ in $disjoint$ and 34.9$\%$ in $overlapped$.

\begin{table}[!t]
\renewcommand{\arraystretch}{1.3}
\setlength{\tabcolsep}{0.15em}
\caption{The effect of fake data generated by SegInversion}
\label{tab:fakedata}
\centering
    \begin{tabular}{c|ccc|ccc|ccc}
    \multicolumn{1}{c}{} & \multicolumn{3}{c}{\textbf{19-1}} & \multicolumn{3}{c}{\textbf{15-5}} & \multicolumn{3}{c}{\textbf{15-1}}\\
    \textbf{Method} & \textit{1-19} & \textit{20} & \textit{all}  & \textit{1-15} & \textit{16-20} & \textit{all} & \textit{1-15} & \textit{16-20} & \textit{all} \\
    \hline
    Baseline & 73.2 & 32.8 & 72.2 & 75.1 & 43.1 & 68.3 & 30.1 & 7.8 & 27.5 \\
    +Noise & 70.7 & 34.6 & 70.0 & 75.0 & 42.8 & 68.1 & 30.4 & 7.8 & 27.6 \\
    HRHF & \textbf{76.8} & \textbf{56.3} & \textbf{75.8} & \textbf{76.2} & \textbf{49.2} & \textbf{69.5} & \textbf{72.6} & \textbf{32.2} & \textbf{62.5}\\
    \hline
    \end{tabular}
\end{table}

\begin{table}[!t]
\renewcommand{\arraystretch}{1.3}
\setlength{\tabcolsep}{0.35em}
\caption{Comparison of mIoU using different aggregation method}
\label{tab:saa}
\centering
    \begin{tabular}{c|ccc}
    \hline
    \textbf{Method} & \textit{1-19} & \textit{20} & \textit{all} \\
    \hline
    FT & 5.8 & 12.3 & 6.2 \\
    MiB & 69.6 & 25.6 & 67.4 \\
    \hline
    AVG & 75.1 & 53.2 & 74.9 \\
    MAX & 74.5 & 54.9 & 74.4 \\
    SAA & \textbf{76.8} & \textbf{56.3} & \textbf{75.8} \\
    \hline
    \end{tabular}
\end{table}

\begin{table}[!t]
\renewcommand{\arraystretch}{1.5}
\setlength{\tabcolsep}{0.35em}
\caption{The influence of different $r$ in SAA}
\label{tab:r}
\centering
    \begin{tabular}{c|c|c|c|c|c|c}
    \hline
    $r$ &    0.5     & 1     &   5   & 10    & 20    & Random\\
    \hline
    \textbf{mIoU} & 75.3    & 75.0  &  74.9 & 74.6  & 74.5  &  \textbf{75.8} \\
    \hline
    \end{tabular}
\end{table}

\subsection{Ablation studies}
To verify the rationality of the HRHF, we conduct extensive ablation experiments on the validation set of PASCAL VOC with different settings for HRHF.\\

\noindent\textbf{The effect of fake data generated by SegInversion}
To verify the effectiveness of fake data generated by SegInversion, we conducted experiments on PASCAL VOC settings as shown in Table~\ref{tab:fakedata}. Our HRHF without using fake data is denoted as ``Baseline''. ``+Noise'' means replacing fake data with random noise data in HRHF. Compared with baseline, using random noise only slightly affects the results. The experimental results demonstrate the effectiveness of the fake data generated by SegInversion.

\noindent\textbf{The effect of the SAA}
To synthesize images by previous trained segmentation network using chosen classes rather than detailed segmentation map, we introduce the Scale-Aware Aggregation approach. To verify the effectiveness of the proposed SAA, we conduct experiments on PASCAL VOC with $19$-$1$ setting. In Table~\ref{tab:saa}, ``AVG'' denotes using average aggregation method in SegInversion, seen in section~\ref{SegInversion}, ``MAX'' denotes using max aggregation method in SegInversion. ``AVG'' denotes using the proposed SAA approach. The experimental results demonstrate that the proposed SAA outperforms the other types of aggregation approaches. It is worth noting that the proposed HRHF outperforms the previous state-of-the-art method by a large margin, \eg, MiB~\cite{cermelli2020modeling}, even if the average aggregation method or max aggregation method is used.
\\

\begin{table}[!t]
\renewcommand{\arraystretch}{1.3}
\setlength{\tabcolsep}{0.15em}
\caption{The effect of the HRHF distillation loss}
\label{tab:dist_loss}
\centering
    \begin{tabular}{c|ccc|ccc|ccc}
    \multicolumn{1}{c}{} & \multicolumn{3}{c}{\textbf{19-1}} & \multicolumn{3}{c}{\textbf{15-5}} & \multicolumn{3}{c}{\textbf{15-1}}\\
    \textbf{Method} & \textit{1-19} & \textit{20} & \textit{all}  & \textit{1-15} & \textit{16-20} & \textit{all} & \textit{1-15} & \textit{16-20} & \textit{all} \\
    \hline
    HRHF & \textbf{76.8} & 56.3 & \textbf{75.8} & \textbf{76.2} & 49.2 & \textbf{69.5} & \textbf{72.6} & \textbf{32.2} & \textbf{62.5}\\
    W/O KD & 71.5 & \textbf{57.8} & 70.9 & 71.2 & \textbf{49.3} & 65.7 & 54.7 & 27.7 & 47.9 \\
    \hline
    \end{tabular}
\end{table}

\begin{table}[!t]
\renewcommand{\arraystretch}{1.3}
\setlength{\tabcolsep}{0.35em}
\caption{The effect of the proportion of fake data in a mini-batch}
\label{tab:proportion}
\centering
    \begin{tabular}{c|ccc}
    \hline
    \textbf{Real:Fake} & \textit{1-19} & \textit{20} & \textit{all} \\
    \hline
    1:1 & \textbf{76.8} & 56.3 & \textbf{75.8} \\
    1:2 & 72.7 & 43.7 & 71.3 \\
    2:1 & 75.2 & \textbf{57.4} & 74.3 \\
    \hline
    \end{tabular}
\end{table}

\begin{figure*}[!t]
    \centering
    \includegraphics[width=1.0\linewidth]{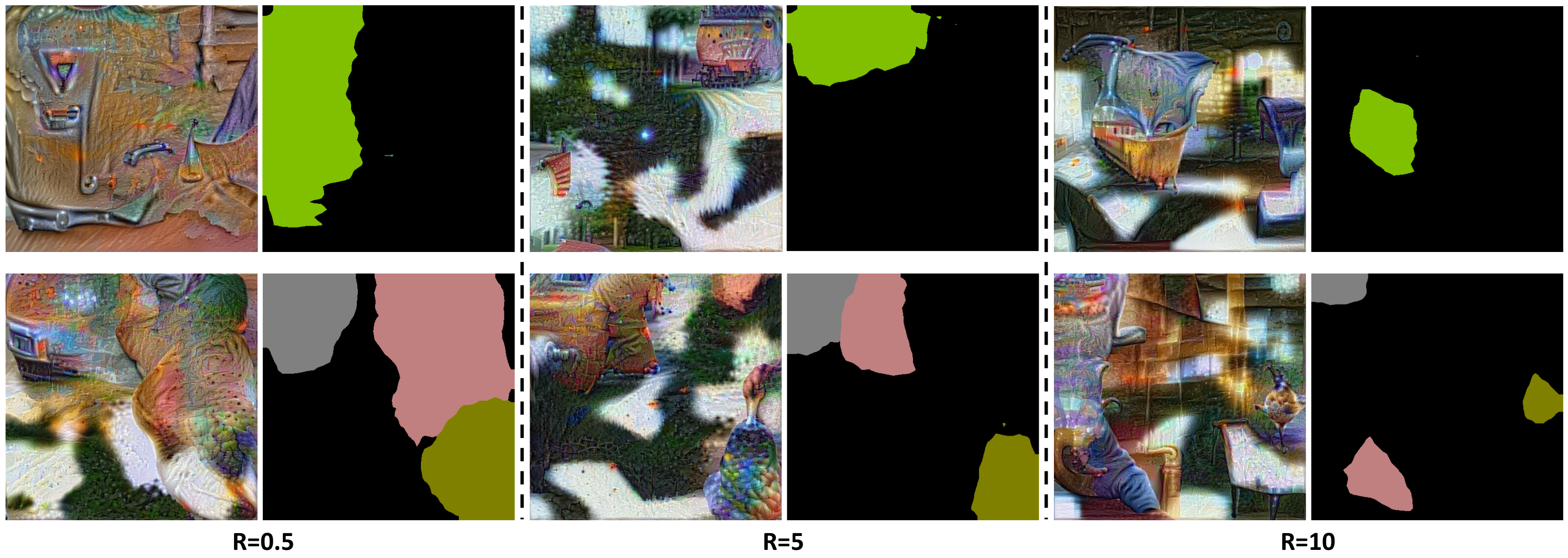}
    \caption{The relationship between the $r$ and the scale of the object. The images in first row contains 2 classes: background and train. The images in second row contains 5 classes: background, person, car and bird.}
    \label{fig:r}
    \vspace{0mm}
\end{figure*}

\begin{figure*}[!t]
    \centering
    \includegraphics[width=1.0\linewidth]{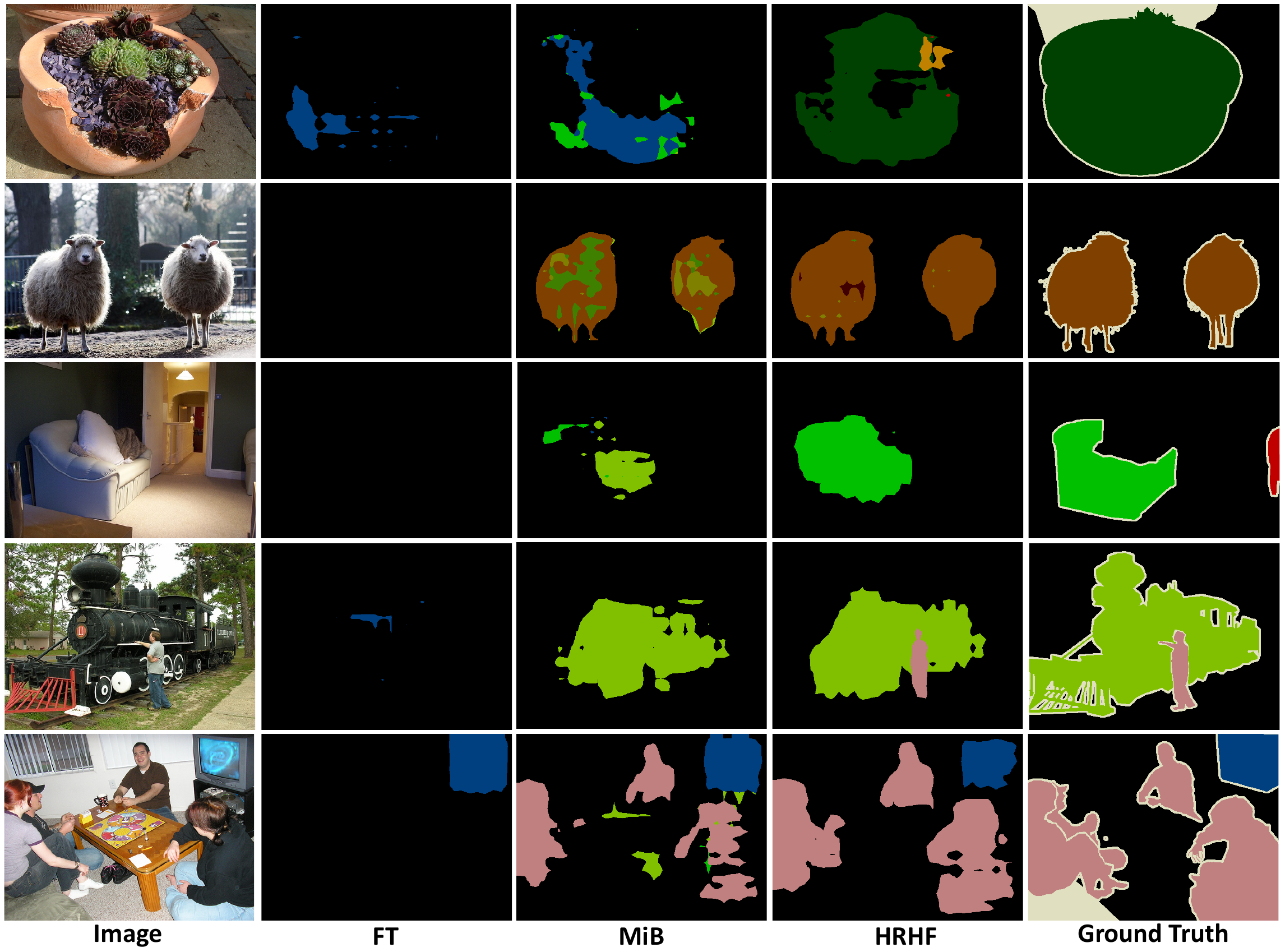}
    \caption{Visualization of several incremental learning methods on PASCAL VOC 2012 validation set. The left column is the input images, the 2nd, 3rd, 4th columns are results of FT, MiB and HRHF respectively. The last column is ground truth. }
    \label{fig:visual_compare}
\end{figure*}

\begin{table*}[!t]
\renewcommand{\arraystretch}{1.3}
\setlength{\tabcolsep}{0.15em}
\caption{Comparison of mIoU on the ADE20K dataset with different methods}
\label{tab:ade}
\centering
    \begin{tabular}{c|ccc||ccccccc||cccc}
    \hline
    {} & \multicolumn{3}{c||}{\textbf{100-50}} & \multicolumn{7}{c||}{\textbf{100-10}} & \multicolumn{4}{c}{\textbf{50-50}} \\
    \textbf{Method} & \textit{1-100} & \textit{101-150} & \textit{all} & \textit{1-100} & \textit{101-110} & \textit{111-120}  & \textit{121-130}  & \textit{131-140}  & \textit{141-150}  & \textit{all} & \textit{1-50} & \textit{51-100} & \textit{101-150} & \textit{all}\\
    \hline
    FT & 0.0 & 24.9 & 8.3 & 0.0 & 0.0 & 0.0 & 0.0 & 0.0 & 16.6 & 1.1 & 0.0 & 0.0 & 22.0 & 7.3\\
    LwF\cite{li2017learning} & 21.1 & 25.6 & 22.6 & 0.1 & 0.0 & 0.4 & 2.6 & 4.6 & 16.9 & 1.7 & 5.7 & 12.9 & 22.8 & 13.9\\
    LwF-MC\cite{rebuffi2017icarl} & 34.2 & 10.5 & 26.3 & 18.7 & 2.5 & 8.7 & 4.1 & 6.5 & 5.1 & 14.3 & 27.8 & 7.0 & 10.4 & 15.1\\
    ILT\cite{michieli2019incremental} & 22.9 & 18.9 & 21.6 & 0.3 & 0.0 & 1.0 & 2.1 & 4.6 & 10.7 & 1.4 & 8.4 & 9.7 & 14.3 & 10.8\\
    MiB\cite{cermelli2020modeling} & 37.9 & 27.9 & 34.6 & 31.8 & 10.4 & 14.8 & 12.8 & 13.6 & \textbf{18.7} & 25.9 & 35.5 & 22.2 & 23.6 & 27.0\\
    HRHF & \textbf{39.8} & \textbf{28.0} & \textbf{35.8} & \textbf{37.0} & \textbf{14.2} & \textbf{19.5} & \textbf{19.4} & \textbf{14.8} & 15.3 & \textbf{30.2} & \textbf{42.5} & \textbf{28.3} & \textbf{25.3} & \textbf{33.3}\\
    \hline
    Joint & 44.3 & 28.2 & 38.9 & 44.3& 26.1 & 42.8 & 26.7 & 28.1 & 17.3 & 38.9 & 51.1 & 38.3 & 28.2 & 38.9\\
    \hline
    \end{tabular}
    \vspace{-4mm}
\end{table*}

\noindent\textbf{The effect of the random $r$ in SAA} To further prove the effectiveness of the hyper-parameter $r$ in SAA. We explore the influence of $r$ with different values. As shown in Table~\ref{tab:r}, the results are evaluated on PASCAL VOC 2012 with $19$-$1$ setting. In fake data generation step, a $r$ in SAA is set to \{0.5, 1, 5, 10, 20\}. ``random'' donates that the $r$ is randomly chosen from a set \{0.5, 1, 5, 10, 20\} for each random input. The high $r$ values implies having an effect similar to the max aggregation and the very low values will have an effect similar to the averaging aggregation. 
The experimental results demonstrate that the results decrease with the increasing of $r$. And, randomly selecting $r$ could achieve better results than the fixed $r$. We argue that the $r$ could control the scale of the object and help increase diversity in fake data. As shown in Figure~\ref{fig:r}, we keep the input noise and model fixed, then set $r$ to different values. With the increasing of $r$, the scale of object decreases. Meanwhile, we try to choose different number of categories as the target. The visualization shows the conclusion is consistent.
\\

\noindent\textbf{The effect of the HRHF distillation loss}
To prove the effectiveness of the proposed distillation framework, we conduct the experiments on PASCAL VOC 2012 with $19$-$1$ setting. The ``HRHF'' donates the complete distillation framework with the segmentation loss and the distillation loss. ``W/O KD'' donates HRHF without the distillation loss. As shown in Table~\ref{tab:dist_loss}, the ``HRHF'' obtains better performance than ``W/O KD'' on the old classes and overall results under all settings. After several update steps under the $15$-$1$ setting, the ``HRHF'' also obtains better performance on the new classes. The experimental results demonstrate that the distillation loss could effectively preserve previous knowledge of old classes.

\noindent\textbf{The effect of the proportion of fake data in a mini-batch}
It is also important to manage the proportion of fake data in a mini-batch. As shown in Table~\ref{tab:proportion}, the experimental results demonstrate HRHF achieves best performance when the ratio of real data to fake data is 1:1. 

\noindent\textbf{Visualization results of several incremental learning methods.}
To get an intuitive understanding of our HRHF, we visualize the results of several incremental learning methods as shown in Figure~\ref{fig:visual_compare}. The experiments are conducted under the $19$-$1$ setting. The $tv$-$monitor$ is the new class, which is marked in blue color. As we can see, the image in the last row contains a monitor, all incremental learning methods could segment the monitor correctly. However, FT and MiB have the bad performance on the old classes, for example, the plants in the first row, person in the third row are misclassified. The visualization shows the proposed HRHF could not only adapts the old network to the new classes, but also have the good performance on the old classes.

\subsection{ADE20K}

ADE20K~\cite{zhou2017scene} is a large-scale dataset that contains images from 150 classes. Following~\cite{cermelli2020modeling}, we use the same setup and create the incremental datasets by splitting the whole dataset into disjoint image sets. In each step it provides annotations only for novel classes while other classes are labeled as background in the ground truth. In Table~\ref{tab:ade} we report our results which still preserve a good performance in this more difficult benchmark for incremental learning of semantic segmentation.

\noindent\textbf{Single-step addition of 50 classes ($100$-$50$).} In this experiment, we initially train the network on 100 classes in the first step and add the remaining 50 all at once. MiB has already achieved good performance, especially in step $101$-$150$, our method succeeds to further improve the overall performance by 1.2$\%$. 

\noindent\textbf{Multi-step addition of 50 classes ($100$-$10$).} Similar to the previous experiment, but instead of adding remaining 50 classes in the second step, we add them 10 by 10 in 5 incremental steps. This is a relatively difficult setting, in which almost all methods performs badly except MiB. Our method still outperforms it by 4.3$\%$. 

\noindent\textbf{Three steps of 50 classes ($50$-$50$).} Finally we analyze the performance on three sequential steps of 50 classes. Our method achieves the best result and improves performance by 6.3$\%$ in total.

\section{Conclusion and future work} \label{Conclusion}
We have studied the incremental class learning problem for semantic segmentation, analyzing the realistic scenario where both old images and annotations are not available, leading to the catastrophic forgetting problem. We address this issue by proposing a new Half-Real and Half-Fake distillation framework which uses real data of new classes and fake data of old classes, effectively learning new classes without deteriorating its ability to recognize old ones. Meanwhile, to generate fake data using the previous trained segmentation model, we proposed SegInversion with a new Scale-Aware aggregation approach to “invert” a trained segmentation network to synthesize class-conditional input images starting from random noise. Experimental Results show that our approach outperforms the other state-of-the-art incremental learning methods. 

{\small
\bibliographystyle{ieee_fullname}
\bibliography{egbib}

\begin{thebibliography}{10}\itemsep=-1pt

\bibitem{azuma1997survey}
Ronald~T Azuma.
\newblock A survey of augmented reality.
\newblock {\em Presence: Teleoperators \& Virtual Environments}, 6(4):355--385,
  1997.

\bibitem{bhardwaj2019dream}
Kartikeya Bhardwaj, Naveen Suda, and Radu Marculescu.
\newblock Dream distillation: A data-independent model compression framework.
\newblock {\em arXiv preprint arXiv:1905.07072}, 2019.

\bibitem{caccia2020online}
Massimo Caccia, Pau Rodriguez, Oleksiy Ostapenko, Fabrice Normandin, Min Lin,
  Lucas Page-Caccia, Issam~Hadj Laradji, Irina Rish, Alexandre Lacoste, David
  V{\'a}zquez, et~al.
\newblock Online fast adaptation and knowledge accumulation (osaka): a new
  approach to continual learning.
\newblock {\em Advances in Neural Information Processing Systems}, 33, 2020.

\bibitem{castro2018end}
Francisco~M Castro, Manuel~J Mar{\'\i}n-Jim{\'e}nez, Nicol{\'a}s Guil, Cordelia
  Schmid, and Karteek Alahari.
\newblock End-to-end incremental learning.
\newblock In {\em Proceedings of the European conference on computer vision
  (ECCV)}, pages 233--248, 2018.

\bibitem{cermelli2020modeling}
Fabio Cermelli, Massimiliano Mancini, Samuel~Rota Bulo, Elisa Ricci, and
  Barbara Caputo.
\newblock Modeling the background for incremental learning in semantic
  segmentation.
\newblock In {\em Proceedings of the IEEE/CVF Conference on Computer Vision and
  Pattern Recognition}, pages 9233--9242, 2020.

\bibitem{chaudhry2018riemannian}
Arslan Chaudhry, Puneet~K Dokania, Thalaiyasingam Ajanthan, and Philip~HS Torr.
\newblock Riemannian walk for incremental learning: Understanding forgetting
  and intransigence.
\newblock In {\em Proceedings of the European Conference on Computer Vision
  (ECCV)}, pages 532--547, 2018.

\bibitem{chen2017rethinking}
Liang-Chieh Chen, George Papandreou, Florian Schroff, and Hartwig Adam.
\newblock Rethinking atrous convolution for semantic image segmentation.
\newblock {\em arXiv preprint arXiv:1706.05587}, 2017.

\bibitem{chen2018encoder}
Liang-Chieh Chen, Yukun Zhu, George Papandreou, Florian Schroff, and Hartwig
  Adam.
\newblock Encoder-decoder with atrous separable convolution for semantic image
  segmentation.
\newblock In {\em Proceedings of the European conference on computer vision
  (ECCV)}, pages 801--818, 2018.

\bibitem{de2019continual}
Matthias De~Lange, Rahaf Aljundi, Marc Masana, Sarah Parisot, Xu Jia, Ales
  Leonardis, Gregory Slabaugh, and Tinne Tuytelaars.
\newblock Continual learning: A comparative study on how to defy forgetting in
  classification tasks.
\newblock {\em arXiv preprint arXiv:1909.08383}, 2(6), 2019.

\bibitem{dhar2019learning}
Prithviraj Dhar, Rajat~Vikram Singh, Kuan-Chuan Peng, Ziyan Wu, and Rama
  Chellappa.
\newblock Learning without memorizing.
\newblock In {\em Proceedings of the IEEE Conference on Computer Vision and
  Pattern Recognition}, pages 5138--5146, 2019.

\bibitem{ebrahimi2020adversarial}
Sayna Ebrahimi, Franziska Meier, Roberto Calandra, Trevor Darrell, and Marcus
  Rohrbach.
\newblock Adversarial continual learning.
\newblock {\em arXiv preprint arXiv:2003.09553}, 2020.

\bibitem{everingham2010pascal}
Mark Everingham, Luc Van~Gool, Christopher~KI Williams, John Winn, and Andrew
  Zisserman.
\newblock The pascal visual object classes (voc) challenge.
\newblock {\em International journal of computer vision}, 88(2):303--338, 2010.

\bibitem{fredrikson2015model}
Matt Fredrikson, Somesh Jha, and Thomas Ristenpart.
\newblock Model inversion attacks that exploit confidence information and basic
  countermeasures.
\newblock In {\em Proceedings of the 22nd ACM SIGSAC Conference on Computer and
  Communications Security}, pages 1322--1333, 2015.

\bibitem{fritsch2013new}
Jannik Fritsch, Tobias Kuehnl, and Andreas Geiger.
\newblock A new performance measure and evaluation benchmark for road detection
  algorithms.
\newblock In {\em 16th International IEEE Conference on Intelligent
  Transportation Systems (ITSC 2013)}, pages 1693--1700. IEEE, 2013.

\bibitem{goodfellow2014generative}
Ian Goodfellow, Jean Pouget-Abadie, Mehdi Mirza, Bing Xu, David Warde-Farley,
  Sherjil Ozair, Aaron Courville, and Yoshua Bengio.
\newblock Generative adversarial nets.
\newblock In {\em Advances in neural information processing systems}, pages
  2672--2680, 2014.

\bibitem{he2016deep}
Kaiming He, Xiangyu Zhang, Shaoqing Ren, and Jian Sun.
\newblock Deep residual learning for image recognition.
\newblock In {\em Proceedings of the IEEE conference on computer vision and
  pattern recognition}, pages 770--778, 2016.

\bibitem{hinton2015distilling}
Geoffrey Hinton, Oriol Vinyals, and Jeff Dean.
\newblock Distilling the knowledge in a neural network.
\newblock {\em arXiv preprint arXiv:1503.02531}, 2015.

\bibitem{hou2019learning}
Saihui Hou, Xinyu Pan, Chen~Change Loy, Zilei Wang, and Dahua Lin.
\newblock Learning a unified classifier incrementally via rebalancing.
\newblock In {\em Proceedings of the IEEE Conference on Computer Vision and
  Pattern Recognition}, pages 831--839, 2019.

\bibitem{huang2019ccnet}
Zilong Huang, Xinggang Wang, Lichao Huang, Chang Huang, Yunchao Wei, and Wenyu
  Liu.
\newblock Ccnet: Criss-cross attention for semantic segmentation.
\newblock In {\em Proceedings of the IEEE International Conference on Computer
  Vision}, pages 603--612, 2019.

\bibitem{huang2020alignseg}
Zilong Huang, Yunchao Wei, Xinggang Wang, Honghui Shi, Wenyu Liu, and Thomas~S
  Huang.
\newblock Alignseg: Feature-aligned segmentation networks.
\newblock {\em arXiv preprint arXiv:2003.00872}, 2020.

\bibitem{kirkpatrick2017overcoming}
James Kirkpatrick, Razvan Pascanu, Neil Rabinowitz, Joel Veness, Guillaume
  Desjardins, Andrei~A Rusu, Kieran Milan, John Quan, Tiago Ramalho, Agnieszka
  Grabska-Barwinska, et~al.
\newblock Overcoming catastrophic forgetting in neural networks.
\newblock {\em Proceedings of the national academy of sciences},
  114(13):3521--3526, 2017.

\bibitem{li2017learning}
Zhizhong Li and Derek Hoiem.
\newblock Learning without forgetting.
\newblock {\em IEEE transactions on pattern analysis and machine intelligence},
  40(12):2935--2947, 2017.

\bibitem{long2015fully}
Jonathan Long, Evan Shelhamer, and Trevor Darrell.
\newblock Fully convolutional networks for semantic segmentation.
\newblock In {\em CVPR}, pages 3431--3440, 2015.

\bibitem{mallya2018piggyback}
Arun Mallya, Dillon Davis, and Svetlana Lazebnik.
\newblock Piggyback: Adapting a single network to multiple tasks by learning to
  mask weights.
\newblock In {\em Proceedings of the European Conference on Computer Vision
  (ECCV)}, pages 67--82, 2018.

\bibitem{mallya2018packnet}
Arun Mallya and Svetlana Lazebnik.
\newblock Packnet: Adding multiple tasks to a single network by iterative
  pruning.
\newblock In {\em Proceedings of the IEEE Conference on Computer Vision and
  Pattern Recognition}, pages 7765--7773, 2018.

\bibitem{mccloskey1989catastrophic}
Michael McCloskey and Neal~J Cohen.
\newblock Catastrophic interference in connectionist networks: The sequential
  learning problem.
\newblock In {\em Psychology of learning and motivation}, volume~24, pages
  109--165. Elsevier, 1989.

\bibitem{michieli2019incremental}
Umberto Michieli and Pietro Zanuttigh.
\newblock Incremental learning techniques for semantic segmentation.
\newblock In {\em Proceedings of the IEEE International Conference on Computer
  Vision Workshops}, pages 0--0, 2019.

\bibitem{michieli2021continual}
Umberto Michieli and Pietro Zanuttigh.
\newblock Continual semantic segmentation via repulsion-attraction of sparse
  and disentangled latent representations.
\newblock {\em CVPR}, 2021.

\bibitem{mordvintsev2015inceptionism}
Alexander Mordvintsev, Christopher Olah, and Mike Tyka.
\newblock Inceptionism: Going deeper into neural networks.
\newblock 2015.

\bibitem{ostapenko2019learning}
Oleksiy Ostapenko, Mihai Puscas, Tassilo Klein, Patrick Jahnichen, and Moin
  Nabi.
\newblock Learning to remember: A synaptic plasticity driven framework for
  continual learning.
\newblock In {\em Proceedings of the IEEE Conference on Computer Vision and
  Pattern Recognition}, pages 11321--11329, 2019.

\bibitem{ozdemir2018learn}
Firat Ozdemir, Philipp Fuernstahl, and Orcun Goksel.
\newblock Learn the new, keep the old: Extending pretrained models with new
  anatomy and images.
\newblock In {\em International Conference on Medical Image Computing and
  Computer-Assisted Intervention}, pages 361--369. Springer, 2018.

\bibitem{ozdemir2019extending}
Firat Ozdemir and Orcun Goksel.
\newblock Extending pretrained segmentation networks with additional anatomical
  structures.
\newblock {\em International journal of computer assisted radiology and
  surgery}, 14(7):1187--1195, 2019.

\bibitem{pinheiro2015image}
Pedro~O Pinheiro and Ronan Collobert.
\newblock From image-level to pixel-level labeling with convolutional networks.
\newblock In {\em Proceedings of the IEEE conference on computer vision and
  pattern recognition}, pages 1713--1721, 2015.

\bibitem{pohlen2017full}
Tobias Pohlen, Alexander Hermans, Markus Mathias, and Bastian Leibe.
\newblock Full-resolution residual networks for semantic segmentation in street
  scenes.
\newblock In {\em Proceedings of the IEEE Conference on Computer Vision and
  Pattern Recognition}, pages 4151--4160, 2017.

\bibitem{rebuffi2017icarl}
Sylvestre-Alvise Rebuffi, Alexander Kolesnikov, Georg Sperl, and Christoph~H
  Lampert.
\newblock icarl: Incremental classifier and representation learning.
\newblock In {\em Proceedings of the IEEE conference on Computer Vision and
  Pattern Recognition}, pages 2001--2010, 2017.

\bibitem{ronneberger2015u}
Olaf Ronneberger, Philipp Fischer, and Thomas Brox.
\newblock U-net: Convolutional networks for biomedical image segmentation.
\newblock In {\em International Conference on Medical image computing and
  computer-assisted intervention}, pages 234--241. Springer, 2015.

\bibitem{rota2018place}
Samuel Rota~Bul{\`o}, Lorenzo Porzi, and Peter Kontschieder.
\newblock In-place activated batchnorm for memory-optimized training of dnns.
\newblock In {\em Proceedings of the IEEE Conference on Computer Vision and
  Pattern Recognition}, pages 5639--5647, 2018.

\bibitem{rusu2016progressive}
Andrei~A Rusu, Neil~C Rabinowitz, Guillaume Desjardins, Hubert Soyer, James
  Kirkpatrick, Koray Kavukcuoglu, Razvan Pascanu, and Raia Hadsell.
\newblock Progressive neural networks.
\newblock {\em arXiv preprint arXiv:1606.04671}, 2016.

\bibitem{shin2017continual}
Hanul Shin, Jung~Kwon Lee, Jaehong Kim, and Jiwon Kim.
\newblock Continual learning with deep generative replay.
\newblock In {\em Advances in Neural Information Processing Systems}, pages
  2990--2999, 2017.

\bibitem{shmelkov2017incremental}
Konstantin Shmelkov, Cordelia Schmid, and Karteek Alahari.
\newblock Incremental learning of object detectors without catastrophic
  forgetting.
\newblock In {\em Proceedings of the IEEE International Conference on Computer
  Vision}, pages 3400--3409, 2017.

\bibitem{tasar2019incremental}
Onur Tasar, Yuliya Tarabalka, and Pierre Alliez.
\newblock Incremental learning for semantic segmentation of large-scale remote
  sensing data.
\newblock {\em IEEE Journal of Selected Topics in Applied Earth Observations
  and Remote Sensing}, 12(9):3524--3537, 2019.

\bibitem{wang2020deep}
Jingdong Wang, Ke Sun, Tianheng Cheng, Borui Jiang, Chaorui Deng, Yang Zhao,
  Dong Liu, Yadong Mu, Mingkui Tan, Xinggang Wang, et~al.
\newblock Deep high-resolution representation learning for visual recognition.
\newblock {\em IEEE transactions on pattern analysis and machine intelligence},
  2020.

\bibitem{wang2018non}
Xiaolong Wang, Ross Girshick, Abhinav Gupta, and Kaiming He.
\newblock Non-local neural networks.
\newblock In {\em CVPR}, pages 7794--7803, 2018.

\bibitem{wu2018memory}
Chenshen Wu, Luis Herranz, Xialei Liu, Joost van~de Weijer, Bogdan Raducanu,
  et~al.
\newblock Memory replay gans: Learning to generate new categories without
  forgetting.
\newblock In {\em Advances in Neural Information Processing Systems}, pages
  5962--5972, 2018.

\bibitem{yin2020dreaming}
Hongxu Yin, Pavlo Molchanov, Jose~M Alvarez, Zhizhong Li, Arun Mallya, Derek
  Hoiem, Niraj~K Jha, and Jan Kautz.
\newblock Dreaming to distill: Data-free knowledge transfer via deepinversion.
\newblock In {\em Proceedings of the IEEE/CVF Conference on Computer Vision and
  Pattern Recognition}, pages 8715--8724, 2020.

\bibitem{yu2018deep}
Fisher Yu, Dequan Wang, Evan Shelhamer, and Trevor Darrell.
\newblock Deep layer aggregation.
\newblock In {\em Proceedings of the IEEE conference on computer vision and
  pattern recognition}, pages 2403--2412, 2018.

\bibitem{yu2020semantic}
Lu Yu, Bartlomiej Twardowski, Xialei Liu, Luis Herranz, Kai Wang, Yongmei
  Cheng, Shangling Jui, and Joost van~de Weijer.
\newblock Semantic drift compensation for class-incremental learning.
\newblock In {\em Proceedings of the IEEE/CVF Conference on Computer Vision and
  Pattern Recognition}, pages 6982--6991, 2020.

\bibitem{zenke2017continual}
Friedemann Zenke, Ben Poole, and Surya Ganguli.
\newblock Continual learning through synaptic intelligence.
\newblock {\em Proceedings of machine learning research}, 70:3987, 2017.

\bibitem{zhang2018context}
Hang Zhang, Kristin Dana, Jianping Shi, Zhongyue Zhang, Xiaogang Wang, Ambrish
  Tyagi, and Amit Agrawal.
\newblock Context encoding for semantic segmentation.
\newblock In {\em CVPR}, 2018.

\bibitem{zhao2017pyramid}
Hengshuang Zhao, Jianping Shi, Xiaojuan Qi, Xiaogang Wang, and Jiaya Jia.
\newblock Pyramid scene parsing network.
\newblock In {\em CVPR}, pages 2881--2890, 2017.

\bibitem{zhou2017scene}
Bolei Zhou, Hang Zhao, Xavier Puig, Sanja Fidler, Adela Barriuso, and Antonio
  Torralba.
\newblock Scene parsing through ade20k dataset.
\newblock In {\em Proceedings of the IEEE conference on computer vision and
  pattern recognition}, pages 633--641, 2017.

\end{thebibliography}
}

\end{document}